# 5 MEANINGFUL HUMAN COMMAND: ADVANCE CONTROL DIRECTIVES AS A METHOD TO ENABLE MORAL AND LEGAL RESPONSIBILITY FOR AUTONOMOUS WEAPONS SYSTEMS

**Author:** S. Kate Devitt

**Author ORCID**: 0000-0002-6075-4969

**Abstract:** 21st century war is increasing in speed, with conventional forces combined with massed use of autonomous systems and human-machine integration. However, a significant challenge is how humans can ensure moral and legal responsibility for systems operating outside of normal temporal parameters. This chapter considers whether humans can stand outside of real time and authorize actions for autonomous systems by the prior establishment of a contract, for actions to occur in a future context particularly in faster than real time or in very slow operations where human consciousness and concentration could not remain well informed. The medical legal precedent found in 'advance care directives' suggests how the time-consuming, deliberative process required for accountability and responsibility of weapons systems may be achievable outside real time captured in an 'advance control directive' (ACD). The chapter proposes 'autonomy command' scaffolded and legitimized through the construction of ACD ahead of the deployment of autonomous systems.

## 5.1 Introduction

> *A legitimate lethal decision process must also meet requirements that the human decision-maker involved in verifying legitimate targets and initiating lethal force against them* **be allowed sufficient time** *to be deliberative, be suitably trained and well informed, and be held accountable and responsible* (Asaro, 2012)

# Meaningful Human Command: Advance Control Directives as a Method to Enable Moral and Legal Responsibility for Autonomous weapons systems

Autonomous systems are attractive for militaries in part because of their capacity to increase mass and operational effectiveness in accordance with human intent and values. Autonomous systems can operate in real time, but may be particularly effective when operating in different time scales to human cognition, such as in high tempo and in slow tempo operations. However, a significant challenge is how humans can ensure moral and legal responsibility for systems operating outside of normal temporal parameters. Paul Ricoeur (1980) emphasized that narrative in time mediates our experience, our interpretation and shared social meanings. Because meaning is constructed through narrative (Maan & Cobaugh, 2018), an evaluation of 'meaningful human control' needs to consider the temporal lens through which the experience of autonomous systems is had. Including the narrative and how meaning is ascribed to the experience, as well as how control of these systems is assessed. A critical factor that affects human intervention and the later judgment of actions is the amount of time available to make decisions—motivating this chapter.

Autonomy reduces the fidelity and usefulness of human participation at an event when the actions of autonomous systems are responding to stimuli faster than a human could comprehend or react to in the time of action (e.g. autonomous countermeasure weapons firing against a hostile missile attack, autonomous cyber offensive when a threat threshold is met). Autonomy reduces the usefulness of human attention and skills during an event when actions of autonomous systems are hibernating, infrequently acting or behaving in extremely low-risk ways (e.g. perceiving their environment, traversing a few knots voyaging in very deep ocean). The narrative told by humans through their experience of the autonomous system may not be best told second by second, but in meaningful segments, constructed by the rippling impacts of their decision making, impacts in the world and effects on themselves. Just as the brain is less active in a monotonous environment, humans can better understand and interpret autonomous system behaviours through cognitively graspable variation and change. This is exemplified by the use of dynamic light and sound for sirens on emergency vehicles to ensure humans remain alert and pay attention. Meaningful and optimal human control of autonomous systems ought to ensure human engagement and mental acuity, which may mean changing the tempo of systems in simulation ahead of operations to trigger appropriate human awareness, engagement and understanding.

This proposal aligns with human factors findings on the effect of boredom (Smith, 1981), lack of engagement and arousal on undesirable neurocognitive states (mind wandering, effort

S.K. Devitt

withdrawal, perseveration, inattentional phenomena) (Dehais et al., 2020). That being said, human factors simulation research has largely studied human attention and action in real time (Hancock et al., 2008). So, solutions recommend changing the user interface, dynamic task allocation and neuro-adaptation of the user (Dehais et al., 2020), rather than changing the tempo of operations to meet human cognitive preferences.

Is it possible, then, for humans to stand outside of real time and authorize actions for autonomous systems by the prior establishment of a contract, for actions to occur in a future context particularly in faster than real time or in very slow operations where human consciousness and concentration could not remain well informed?

This chapter addresses this question by drawing on legal precedent in healthcare via advance care directives, where humans are asked to consider what they wish to be done medically upon them when they are no longer capable of making decisions for themselves. This medical precedent, it will be argued, can shed light on how the time-consuming, deliberative process required for accountability and responsibility of weapons systems may be achievable outside real time captured in an 'advance control directive' (ACD). For decision makers to be accountable, they would need embark on 'advance control planning' (ACP) in advance of an ACD; working with systems extensively in development and training including simulation, exercises, trials and in theatre gaining real knowledge of how autonomous systems operate to be confident that a) legitimate targets can be identified by the system; and b) humans will be held accountable and responsible for their deployment.

## 5.2 Definitions

First for some definitions. When I refer to 'autonomous weapons system' (AWS), I draw on the US DoD 3000.09 (Department of Defense, 2023):

> *AWS: A weapons system that, once activated, can select and engage targets without further intervention by an operator. This includes, but is not limited to, operator-supervised autonomous weapon systems that are designed to allow operators to override operation of the weapon system, but can select and engage targets without further operator input after activation.*

Thus it is definitional that there are some actions taken by an autonomous system without human intervention for some time $t_m$ to $t_n$. For non-weaponized systems 'autonomous systems' (AS) I abstract from the US DoD 3000.09 to mean:

> *AS: A system that, once activated, can act without further intervention by an operator. This includes, but is not limited to, operator-supervised autonomous systems that are designed to allow operators to override operation of the system, but can act without further operator input after activation.*

Note that a non-weaponized AS may still act in a problematic way or cause harms. For example, a commander may authorize a fleet of uncrewed underwater vehicles (UUVs) to traverse a segment of ocean, which might accidently damage protected reef (Gu et al., 2021) or stray into territorial waters of a foreign state (McKenzie, 2020, 2021).

When I refer to responsibility, I mean command responsibility (Mettraux, 2009), which requires that humans meet the criteria for moral responsibility and legal responsibility which I discuss in more depth in the Command section below. But briefly, the basic conditions for moral responsibility and blameworthiness requires a human to have autonomy, knowledge and capability to make a decision and can reasonably forsee the consequences of that decision (Talbert, 2016). A person is not morally responsible if they are coerced into a decision, could not have foreseen the outcome of their decisions or were incapable of making the decision due to circumstances outside of their control. Legal responsibility requires an appropriate connection between a human operator and the autonomous system, namely that the human directs the activities of the autonomous system and is accountable for the system's behaviour (Rain Liivoja et al., 2022).

### 5.3 Problems

There are three responsibility problems for humans using autonomous systems: 1) real-time responsibility for many autonomous systems operating in parallel; 2) responsibility for fast-tempo autonomous systems and; 3) responsibility for slow-tempo autonomous systems. I will describe each briefly. Preparing for real-time autonomy with many systems operating in parallel (such as a robot swarm) would need a way to decompose swarm decision-making and parameters of actions ahead of operations to ensure appropriate human comprehension and oversight. For example, swarms could be tested working at fast speeds in simulation with numerous risks and the varied results of tests and analysis shown to humans to help develop guidance for their use[1]. Fast-tempo autonomous systems (e.g. cyber countermeasures) would need to be slowed down in simulation, so that humans can work through each decision including probabilistic reasoning. On the other hand, slow-tempo autonomous systems (e.g. long-range uncrewed undersea surveillance) would speed up decisions shown in simulation so that humans can experience human-speed actions against which to judge systems behaviour including their ability to meet objectives as well as ethicality and legality.

The chapter is organized as follows. First, the chapter identifies the functional speed of autonomous systems as a significant risk to humans achieving meaningful human control of autonomous technologies. The chapter considers three factors: 21st century war trends, advance care directives and command including moral and legal responsibility. The findings from these sections inform the concept of 'autonomy command' indicating the potential similarities and differences between traditional command and command of autonomous systems. Considering autonomy command, a process to build ACD is postulated and a case study of Australia is given to provide a national context. Finally, the chapter considers the role of artificial intelligence in future autonomy command and the interplay of human decision makers working increasingly in digital simulations to determine appropriate courses of action. The chapter concludes by reinforcing the challenge for militaries seeking to maintain moral and legal responsibility for autonomous systems and the potential value of the preparation of an ACD to govern their use.

This chapter only proposes a method by which to consider developing such a contractual framework. It falls short of stipulating specific legal or ethical requirements, prohibitions, operational limits or an evaluation of risks that would be required for each instantiation of an ACD for deployment of an autonomous weapons system.

### 5.4 Meaningful Human Control

In accordance with the definition of autonomous systems above, an autonomous agent (human or artificial) will sometimes choose its own actions without human intervention.

Meaningful Human Command: Advance Control Directives as a Method to Enable Moral and Legal Responsibility for Autonomous weapons systems

Autonomous agents will be influenced by internal and external factors (such as programming and environment) and may be restricted or bounded; but will, for some period of time within some spactial boundaries, be able to determine their own actions. Can such systems be under meaningful human control? Or if not control, perhaps remain under meaningful human command? Issues of control certainly come up on the context of future weapons.

On the one hand, the International Committee of the Red Cross (ICRC) argue that there must be effective human supervision, and timely intervention and deactivation of autonomous weapons systems (International Committee of the Red Cross, 2021). They also argue that there should be a ban on autonomous weapons systems that select and apply force to human targets (International Committee of the Red Cross, 2021, 2022). The ICRC argue that humans must maintain control of weapons systems, that is, humans need to be involved in both targeting and the application of force.

> *The use of autonomous weapon systems entails risks due to the difficulties in anticipating and limiting their effects. This loss of human control and judgement in the use of force and weapons raises serious concerns from humanitarian, legal and ethical perspectives.* (International Committee of the Red Cross, 2021)

On the other hand, militaries have been controlling weapons systems with software to enable their lawful operation for 40+ years (Department of Defense, 1978). Defence industries now advocate for what they call 'kill chain acceleration'. Kill chain acceleration means that military actions move faster and faster to achieve military objectives, bypassing limits on human cognition[2]. Is it possible to have 'meaningful human control' over these fast tempo actions? Philosophically, the answer is 'it depends'. It depends, in particular, on what is meant by 'meaningful' and 'control'.

The term 'meaningful human control' is thus contentious with regards to the governance requirements of AI-enabled, robotic and autonomous (RAS-AI) weapons systems. This is partly because determining the parameters of 'meaningful' is subjective, complex, pluralistic and ambiguous. Some might argue why the term 'meaningful' is required, rather than simply requiring 'control' ?

Perhaps the logbook of autonomous platforms can aggregate meaningless time into small chunks between meaningful affairs, removing boring, uneventful events and drawing human attention when they detect change or ambiguity. This is how the Ocius Bluebottle

# Meaningful Human Command: Advance Control Directives as a Method to Enable Moral and Legal Responsibility for Autonomous weapons systems

(Department of Defence, 2023) traverses the ocean conducting surveillance. Fully autonomous functions are replaced by remote operation when collision avoidance and other human oversight is required[3]. In this way, the 'story' of the actions of an autonomous platform may be best told communication at a pace suitable to human cognition about both internal and external matters:

- internal matters might include incident reports, self-repair, loss of sensors or communications that degrades decision quality, perception of surprising environments,
- external matters might include interactions with foreign objects, application of force, including manipulation (e.g. moving debris, taking samples, holding and repairing cables) engagement (e.g. pushing objects, application of kinetic force), communication and cyber engagement.

However, the question of *meaningful* human control of artificial intelligence (AI) arises particularly in situations where it is feared that artificial agents have agency, autonomy and decision authority in circumstances of ethical or legal risk, where humans should have oversight and capacity to intervene. Sometimes, when humans are considered *not* to be in meaningful control of AI, it means that they are unable to be held morally or legally responsible for decisions made by the system. Conversely, humans are considered to be in meaningful control when they are morally and legally responsible for decisions they make with an AI. There is a lot to debate here and this chapter does not aim to resolve the question of meaningful human control, but instead focusses on responsibility and accountability for the use of autonomous weapons systems which is likely to feed into a robust account of meaningful human control.

One way to frame the debate is to understand meaningful human *control* through meaningful human *command*. Command ensures responsibility and accountability—a key principle of NATO's AI Strategy (NATO, 2021)—of how technologies are deployed without necessarily requiring constant direct control. NATO's AI Strategy is the strongest international commitment to responsible use of AI for military systems to date. This chapter explores meaningful human command, specifically, how 'autonomy command' might be scaffolded through the construction of ACD ahead of the deployment of autonomous systems. But first we need to situate autonomy command in the realities of current and future war and command.

## 5.5 Trends in 21st century war

> *It's like I'm reading a book, and it's a book I deeply love, but I'm reading it slowly now, so the words are really far apart and the spaces between the words are almost infinite. I can still feel you and the words of our story, but it's in this endless space between the words that I'm finding myself now— Samantha AI OS (Jonze, 2013).*

The AI operating system (OS) assistant Samantha in the movie 'Her' evolves to a point where for each second of human experience, she processes information and has experiences orders of magnitude faster and more complex than her human companion and so she and the other AI OS abandon their human companions. While conscious AI is a long way off (Gebru & Mitchell, 2022; Schneider, 2019), new trends in defence technologies include more AI, faster information processing and decision cycles. For example, see the marketing patter from highly successful Silicon Valley defence-industries start-up Anduril regarding their AI-infused decision support product Lattice:

> [Lattice] enables men and women in uniform to *move with machine speed*, unparalleled confidence, and military-grade security by turning data into information, information into decisions, and decisions into actions across tactical and strategic operations (Anduril, 2022) [italics added].

In 'War Transformed', Mick Ryan (2022) highlights seven trends in 21st century war. I highlight three relevant to this chapter below:

T1. Increasing speed of planning, decision making and action.

T2. Large-scale conventional forces combined with the massed use of autonomous systems and extensive influence tools.

T3. Human-machine integration. Autonomous systems will be full partners with human beings in the conduct of military missions.

Ryan argues that these trends require new twenty-first-century warfighting concepts and strategies developed by people experienced in human-machine integration. These concepts must "augment human physical and cognitive capabilities to generate greater mass, more lethal deterrent capabilities, more rapid decision-making, more effective integration, more efficient training and education, and better experimentation and testing" (Ryan, 2022, p. 84).

However, how can T1-3 possibly be achieved by nations seeking to be ethical and law abiding? If T1 is true, and real-time decisions in war are speeding up, but human cognition remains at the same speed—then surely the inevitable outcome of ever-increasing speeds is a loss of human awareness, understanding, agency and autonomy over machines and hence loss of moral and legal responsibility. This point is one of many made by those concerned by introduction of AI and autonomy on the battlefield (Asaro, 2012; Sharkey, 2012, p. 116).

In response, militaries might argue that many ethical decisions can be sped up with training, just as soldiers can practice prioritizing first aid to harmed blue forces in exercises even if it delays achievement of military objectives; and fighter pilots can practice swift and precise responses based on updated, perhaps more nuanced targeting parameters. However, while unconscious, gut-level, heuristic, 'rule of thumb' ('type 1') ethical actions can be sped up (to a point); conscious, slowed-down, reflective and deliberative reasoning and decision making, ('type 2') ethical actions cannot be accelerated (Kahneman, 2011; Stanovich & West, 2000). Indeed, it can be hard to motivate humans to even to apply slower reasoning to decisions in normal time, let alone manage their biases, fallacies and mindset (Hendrickson, 2018) or trying to speed them up in stressful, fatigued and uncertain circumstances (Lieberman et al., 2005).

The distinction between 'type 1' and 'type 2' kinds of reasoning has been widely disseminated in leadership, business and decision-making conversations, and is meant to capture the difference between underlying cognitive processes, although this is controversial and the brain may not in fact, easily be bifurcated in this way (Sorensen, 2016) but also such a distinction may miss other important variables such as emotion (Roeser, 2012) and motivation that affects decision making. In this chapter I ask the question about how the speed and urgency of decisions may affect responsibility and, for the sake of brevity, remain agnostic with regards to the cognitive mechanisms engaged.

So, do militaries have an impossible task to meet the reality of accelerated planning, decision-making and action, keeping moral and legal responsibilities intact? Can responsible militaries formulate a new governance mechanism that allows for systems in T2 to be authorized to act autonomously in high-tempo and slow-tempo operations within an acceptable ethical and legal framework? To capture 'type 2' thinking, such a framework would require a set of responsible, authority-bearing humans to select a set of appropriate future actions for autonomous systems outside of temporal pressures and in consultation with experts.

It is in the context of advanced 'type 2' thinking ahead of time where the model of advance care planning and advance care directives in health care can be considered a model. Type 2 thinking in advance for command requires planning and the formulation of directives for the future actions of autonomous systems with legal and ethical responsibilities intact and accountability lines clearly drawn.

The chapter takes as a premise that Human-AI trust in military operations can be contractual, and that this contract must be explicit. Contracts prove a useful mechanism for agent governance because the parameters and limits of trust can be specified explicitly. To trust the agent is to believe that a set of contracts will be upheld. Contracts implies an obligation by both the AI and the AI developer to carry out a prior or expected agreement (Jacovi et al., 2021). In this way, incidents when they occur, are to some degree explainable; evaluable and given due precautions to limit reputational damage. If unreasonable actions are taken, then the contracts established can be scrutinized in and after incident legal review. Additionally, the actions of AI may not be fully explainable by recourse to a contract, in the case of emergent behaviours. An open question is whether autonomous system behaviours can be predictable enough, to meet ethical and legal concerns (International Committee of the Red Cross, 2021).

## 5.6 Advance care directives
*5.6.1 Prelude*

If a contract is possible between humans and autonomous sytems to ensure responsibility, what sort of contract might it be? In the military domain the sort of contract is likely to connect with rules of engagement (ROE) (Hosang, 2020) and other legislative requirements regarding recording military decision making (Devitt et al., 2021, Section 3.5.2). This chapter is not written by a lawyer or legal expert, but a philosopher and cognitive scientist. So, the exploration of contracts from the medical domain is meant to encourage creative thinking in terms of the sort of processes and content that would be relevant, appropriate and useful in a command contract to ensure responsibility. Formulating the appropriate legal instrument and legal mechanisms are some of the next steps if the philosophical premise is encouraging.

If exploration of contracts is a worthwhile philosophical endeavour, then, is there legal precedent for the formation of a contract at a point $t_m$, that is (1) valid to direct the actions of future agents at $t_n$, where (2) the consequences of any action include significant harms, suffering and loss of life and (3) finally the person responsible for those harms is the person making the decision at $t_m$?

The advent of advance care directives in the medical domain are an example of such a contract. The reader is invited to consider whether there might be lessons from the past 40+ years of work in health when it comes to directing autonomous systems and being responsible for their future behaviours.

*5.6.2 Summary*

Advance care planning and advance care directives emerged in the 1970s for patients to resist unwanted treatments (such as tube feeding or cardiopulmonary resuscitation) near the end-of-life (Morrison, 2020). Advance care planning is meant to enable patients to articulate their values and goals and how various treatments would align or not align with these goals. Advance care planning addresses: social and cultural beliefs as well as structural constraints of health and legal systems; interventions from multiple stakeholders; interactive information interventions with knowledgeable experts to manage concerns; repeated conversations and evaluations across different settings, strategies to resolve conflicts and implementation requirements (Jimenez et al., 2018, Table 4; Luckett et al., 2015).

The resultant documentation is meant to enable trusted surrogate decision makers and medical professionals to ground their actions based on value alignment to ensure that patients avoided unwanted treatments. While advance care directives have demonstrated some utility, such as with dementia patients and for elderly patients in New York during the COVID-19 pandemic, there is a wealth of evidence that they fail to be linked to outcomes of importance (Morrison, 2020). Decisions near end of life are not simple, logical or linear and change rapidly with changing clinical conditions. It turns out to be extremely challenging to have the sort of care conversations at a sufficient depth to allow for the permutations of real-life scenarios including prognosis, disease, comorbidities and treatment outcomes. Those who wish to deploy autonomous systems in complex, confusing and unpredictable conflict environments ought to take seriously the evidence from health care that making plans is extremely difficult and outcomes often discordant to what is envisioned or signed off on. That being said, autonomous systems are not human beings and they can be limited. It may be possible that a subset of functions can be planned for that provide acceptable behaviours to humans responsible for their deployment. The autonomous system will not be capable of changing its mind about the agreement or deviating from an authorized set of actions.

One might question whether the medical scenario offers an analogous decision-making environment to a human commander directing autonomous systems. For example, one might argue that human commanders are not putting themselves at risk by signing off on an advance control directive, changing the decision calculus and perhaps their commitment to the process. While that is true, it is not the case that the human commanders put themselves in no risk—particularly if they become morally and legally responsible for the actions of autonomous systems. Even if physical injury is less likely, the potential for moral injury is great for autonomy commanders if the systems they command behave in problematic ways or unintented harms are caused (Phelps & Grossman, 2021).

Arguably the health literature is useful for militaries to learn from—even if the analogy is flawed—because there are factors critical to future health decisions that are relevant to future command. For example, mental capacity and provision of information are used to scrutinize a person's autonomous wish in advance health directives. Human consent for a directive does not consist in merely a 'yes' or a 'no' to the provision of treatments. The process by which the determination is made is privotal to its validity. Misleading or insufficient information that leads the person to choose an action that they would not choose if they had access to better information affects the validity of the directive they establish. If we consider this line of argument to autonomy command, then commanders are not truly autonomous agents in agreeing to a set of future actions if they are not provided the full set of relevant facts and information pertaining to the specific deployment of assets at a future time. A commander who was given partial information to satisfy a specific agenda would not be able to act autonomously sufficient to be morally responsible. The next section considers the command function including the conditions for ethical and legal responsibility.

**5.7 Command**

> *We must remind ourselves that it is simply not possible to construct a model for the art of war that can serve as a scaffolding on which the commander can rely for support at any time. Whenever he has to fall back on his innate talent, he will find himself outside the model and in conflict with it; no matter how versatile the code...talent and genius operate outside the rules, and theory conflicts with this practice—Carl von Clausewitz, On War* (von Clausewitz, Howard, & Paret, 1976)

What is it to command? Clausewitz understood that war intrinsically involves uncertainty, a lack of predictability and unexpected circumstances that evades a model of war and requires human creativity, resilience and problem-solving. Any humans or machines in war must be able to adapt to changing circumstances. Humans must be accountable for who and what they deploy. In this chapter I take as a premise that officers must invest in the deliberate and continuous study of the profession of arms including military history, ethics, strategy and leadership to succeed as tactical commanders in fast-changing 21$^{st}$ century society—see Box 5.1 (Chambers et al., 2017; Ryan, 2022). What does it mean to have moral and legal responsibility in command?

> **Box 5.1 Case study: the role of the commander in Command and Control (C2)**
>
> C2 in the Australian Army. Despite increased investments in personnel, ICT and information collection, the differentiator in the success of C2 is the commander. The commander is the 'X Factor' of C2.

> The experience, talent and expertise of the commander (and to a lesser extent the members of the staff), which manifests as capacity to cut through the fog and friction of war, in order to make appropriate, quick and intuitive judgements, are the primary arbiters of the quality of the performance of the headquarters….[T]hese qualities transcend technology, systems, processes, and the size of the staff. (Chambers et al., 2017, p. 9)
>
> The authors of the above quote recommended that the Australian army invest in a broad range of tactical experiences, mentors and professional education. A critical component of professional education should involve training with technologies in war game simulations where future commander decision making can be evaluated and honed.

*5.7.1 Moral responsibility*

Several philosophers argue that moral responsibility is only possible by human agents and possibly sentient machines. Moral responsibility usually requires 1) autonomy and agency over one's decisions (free will), 2) knowledge of the circumstances (situational awareness), and 3) the capacity to act during events involving ethical risk in accordance with intent (capable action) (Crawford, 2013; Talbert, 2016). As Crawford explores in her book, militaries like to represent the normal behaviour of personnel to be 'good apples', that is soldiers who perform their duties, cognisant of ethical and legal obligations and do their best under incredible duress and information uncertainty. 'Good apples' are contrasted to 'bad apples' and 'mad apples'. Bad apples are soldiers who know what is expected of them and how to behave, but deliberately behave contrary to the dictates of their military education, experience and operational circumstances. Mad apples are soldiers who have lost the ability to keep clear about what they ought to do and how they ought to behave. Mad apples are created through a combination of character traits and environmental factors that reduce agency, knowledge or capacity to act ethically and legally. Bad apples are responsible in a way that mad apples are not. The importance of mental state to culpability is reflected in legal systems where individuals who don't know what is going on or deny agency over their own actions or deny acting in accordance with their own intent are not considered responsible for their actions, but are also judged 'criminally insane' and measures are taken to restrict their freedoms (Australasian Legal Information Institute, 2022)[4]. There may be a fourth category of apple, a 'cooked' apple, where an otherwise sane and rational soldier becomes the 'moral crumple zone' in complex information environments (Devitt, 2023; Elish, 2019). Cooked apples are not responsible for their decisions because they have lost autonomy, situational awareness and capable action due to their diminished role in a decision system, rather than due to insanity or other mental problems.

*5.7.2 Autonomy*

Achieving free will or the autonomy of command is critical for the thesis of this chapter to work. If commanders do not have autonomy when they deploy autonomous assets, then they cannot bear moral or legal responsibility for their actions. Kant conceived the will as "the property a will has of being a law to itself" (independently of any property of the objections of volition) (Kant, 2011). What does this mean? Well, it has been interpreted to mean that a person with autonomy is both a rational agent and has their own standards against which they must conform—it does not mean that they get to do whatever they like. The notion of an individual having their own standard is familiar in assessments of proportionate response in just war principles and embedded in international humanitarian law (IHL). Commanders have a legal obligation to determine for themselves what is proportionate in the circumstance of application of force—it is an individual assessment rather than compliance with an objective measure. In fact, international case law developed an objective standard (or at least a standard that "aspires" to objective), that of "reasonable military commander" (Hasson & Slama, 2023; Henderson & Reece, 2018; United Nations International Criminal Tribunal for the former Yugoslavia, 2000, para. 50)[5].

Kant saw human reason as a 'guiding compass' helping to steer an action between good and evil in alignment with duty, independent of personal desires. Kant also saw autonomy as the foundation of human dignity and the source of all morality. That is to say, a commander cannot be morally or legally responsible for decisions made in the absence of autonomy. If a commander is coerced or manipulated or suffer automation bias in their application of force, then they have abdicated their duty as rational and moral agents. For Kant, a person is autonomous if they are rational and unaffected by fear of punishment or desires, free from internal and external pressures. Some might suppose Kant's requirements of autonomy are too onerous and require too much from the individual.

Meaningful Human Command: Advance Control Directives as a Method to Enable Moral and Legal Responsibility for Autonomous weapons systems

John Stuart Mill offered an alternate account. For Mill it is important for the human to be able to decide: "He must use observation to see, reasoning and judgement to foresee, activity to gather materials for decision, discrimination to decide, and when he has decided, firmness and self-control to hold to his deliberate decision" (Mill, 1860).

Mill valued individuality and spontaneity but also expected decision-makers to focus all their facilities to decide the best course of action to pursue. While individual action (liberty) was important, Mill could see that persuasion and advice could be consistent with an autonomous decision, so long as the individual ultimately applied their reasons to the decision so that the advice did not override the individual's reason. We can see the potential threat to rationality and thus autonomy in AI decision support aids.

Interference from others undermines autonomy in the current legally accepted understanding of autonomy unless the person consents to their involvement (Chan, 2018). The degree to which a decision is voluntary is complicated when the decision-maker is embedded within complex and persuasive information environments. Particularly when environments seem to offer rational alternatives, but do so paternalistically (treating the decision-maker as a child, rather than an intellectual equal (Tsai, 2014)). This relationship is well demonstrated in the movie *Wall-E* when the Captain struggles to determine his own mind when given persuasive arguments from the ship's AI (Stanton, 2008). Proponents of nudging will argue that corrective manipulation is ethically defensible when humans have total choice and the consequences for rearranging information will reduce overall harms. For example, in 2016, President Obama issued an executive order "Behavioral Science Insights Policy Directive" (EO 13707 2015)(Lepenies & Malecka, 2018), and one of the consequences was making healthcare 'opt out' rather than 'opt in' because this would insure thousands more vulnerable Americans. However, significant critique of nudging makes it a contentious theory to adopt (Kuyer & Gordijn, 2023). More concerning is when decisions with ethical or legal ramifications are made through a box ticking ritual rather than inviting serious consideration (Chan, 2018).

In cognitive science, 'command' is tied to volition and agency (Haggard, 2017; Haggard et al., 2002). One is in command of own's own body to the degree that one is able to direct it to do as one wishes as opposed to involuntary actions such as reflexes or pure responses to external stimuli (e.g. withdrawing one's hand from a hot surface). Agency is a curious mental state. For example, a human has agency over planting a garden and feels responsible for its success six months later at harvest time. This ability of humans to not merely feel agency at the time of action, but over periods of time is relevant to the challenges of autonomy in high tempo and slow tempo operations. The fact that humans can and do feel agency and responsibility for actions over differing temporal periods becomes a premise for this chapter's normative framework for responsibility of autonomous actions existing at different time scales.

If humans get used to autonomous systems doing very little, it will feel like 'grass growing' or 'paint drying'. Human attention will wane, and they lose their sense of agency. It is better to report on these slow systems via faster narrative time to human overseers. If humans get used to autonomous systems making decisions all the time faster than they can keep up, they will also loose a sense of agency and feel out of their depth, that they don't know what is going on and that they are powerless to intervene. They may avoid intervention, even if things seem to be going badly, to avoid being blamed for their actions.

Prioritizing autonomy in commanders does not prevent unethical acts or guarantee moral responsibility. However, prioritizing command autonomy will affect design choices when developing information and decision interfaces between humans and machines.

*5.7.3 Knowledge*

In the medical literature, 'informed consent' is a key concept. In advance care planning, patients must be informed insofar as possible with regard to likely outcomes and consequences of their decisions. So too commanders must have access to experts to provide analyses of possible outcomes of decisions and their likelihoods. Situational awareness is a common term in military contexts, and it must be considered for fast-tempo and slow-tempo operations from the perspective of advance situational awareness—being able to imagine and map out future scenarios in sufficient detail to be responsible for them when they occur even if the commander is not in contact in real time (Endsley, 2023).

In the medical field, advance care directives hold more weight legally when patients are highly informed, able to understand the nuances of their potential treatments and had very good and sustained conversations with expert medical teams (Chan, 2018). Patients with a record of discussing end-of-life care in an ad hoc way or with only a single conversation are likely to be viewed as having less credence. Similarly in the evaluation of acts of war, those that are prepared with care, nuance and with expert advice, are more likely to be considered as being conducted responsibly and meaningfully.

*5.7.4 Capability*

Commanders must be mentally capable of the job being asked of them. The kinds of capabilities required to effectively command high-tempo and slow-tempo operations will involve familiar and traditional command skills (such as political, strategic, operational and tactical knowledge; legal, safety and ethical reasoning) with the addition of specialized abilities to command AI and autonomous systems, such as:

1. an ability to reason in complex hypothetical scenarios,
2. knowing the capabilities of AI and autonomy platforms including their reasoning, key programmes, logic, 'plays', functional breakdown, operational redundancies, manoeuvres and halting conditions,
3. ability to communicate with machines including conveying command intent and being able to listen to machine status and reports including measures of confidence,
4. understanding human and machine causal chains of action and reaction,
5. probabilistic reasoning.

The skills and knowledge of an autonomy commander will need to be assessed in training and operationally in after-action reviews. Commanders will need to practice coordination of decision making with autonomous platforms in normal time (aka either sped up or slowed down from their operational tempo) to learn and practice coordinated actions such that they can be confident in writing an ACD. The ability to work in simulation, with time to review and reflect on the outcomes and consequences of decisions in hypothetical scenarios will be how a commander builds legitimacy in programming their ACD and the confidence of the state in ratifying such agreements.

It will be up to each nation to decide how much training is required of commanders. If suitable tests could be established, then commander behaviours could be benchmarked against these. Such tests could measure commander cognitive capabilities such as in probabilistic reasoning, memory and decision making as well as scenario-specific capabilities.

Advance control planning will be as much about what systems will not do, as about what they will do. Commanders need to take responsibility for inaction, as well as action, unethical acts by omission as much as commission. For example, if an autonomous vessel is nearby a stranded vessel carrying Australian citizens and could supply assistance, should it? Should it reveal its location to reduce civilian suffering? Will inaction cost the commander?!

*5.7.5 Legal responsibility*

What is the legal meaning of command? In this chapter I will orient discussion to align with modern militaries and the history of military command as represented in international legal frameworks. Command is an act of exercising authority to implement orders or directives and comes with responsibility for those decisions. Military units, whether human or machine, must be under command of a member of the armed forces legally recognized to be a party to a conflict (R. Liivoja et al., 2022). To exercize authority, commanders must process information to make decisions (van Creveld, 1985). They must obey orders but do have discretion to interpret those orders and deviate from a plan when required. This is important for orders, but also for planning[6]. Moltke wrote that "Kein Operationsplan reicht mit einiger Sicherheit über das erste Zusammentreffen mit der feindlichen Hauptmacht hinaus" (Hughes, 1993), and so commanders need to be agile not only in their execution but also in their planning. In the battle of Goose Green in the Falklands War, use of a phased detailed plan (as opposed to Auftragstaktik, or "mission orders") nearly led to disaster for the British forces (Fitz-Gibbon, 2006). In this case commanders need to be agile but so do the autonomous systems. Commanders have autonomy and agency to exercise human discretion and maintain responsibility for this process and result.

What does the law say about commanding autonomous systems that may be physically and temporally separated from technologies? As a new weapon, means or method of warfare, autonomous weapons systems trigger an Article 36 review to determine whether their use is prohibited by international law (International Committee of the Red Cross, 2006). However, international legal frameworks do not specifically preclude automation or autonomy in military platforms, which means determining the limits of their use is open to interpretation[7].

Command requirements are such that systems must fulfil the intent of the commander such that a legal link can be made between the commander and the intention of the deploying state.

As R. Liivoja et al. (2022) indicate,

> …States do and will increasingly aim to use systems where the person responsible for the use of force is physically and temporally separated from the device which delivers the desired effect. While it might not hold for every automated and autonomous device or every circumstance, where an adequate connection can be shown with a human operator… the device will still be under command as there will be a specific identifiable who is directing its activities and accountable for its behaviour. (R. Liivoja et al., 2022, p.32)

The following sections assume that States will wish to develop autonomous systems that are at times physically and temporally separated and wish to demonstrate an adequate connection between a human operator and these devices during these separations.

## 5.8 Autonomy command

Command is changing with the rise of AI and robotics and autonomous systems that incorporate AI to determine their actions. Ryan (2022) argues that "AI will change the balance of power in tactical military endeavors" (p. 42). Responsible use of AI may increase lethality and reduce risk to the forces deploying it. AI will assist in decision support, and support alignment of tactical goals with higher aims as well as increase the speed of decision making. AI in simulations can assist in tactical planning by rapidly modelling outcomes of many options in development, test and evaluation and preparation as well as during military activities.

Ryan (2022) considers a joint military task force of the future employing hundreds of human personnel to deploy several thousand robotic systems of various sizes, functions and configurations such as swarms and super swarms (systems of swarms). These human-robot teams will need to balance autonomous functions with human responsibility for the actions of robotic assistants fulfilling commander's intent. Ryan asks how robots on the battlefield (Freedberg Jr., 2020) including swarms will be governed. This section proposes a method by which appropriate governance of autonomous systems can be assured in operations particularly where humans cannot reasonably intervene using ACD.

ACD govern the behaviour of autonomous systems that fall outside of communication with humans where humans cannot intervene on their behaviours during a specified temporal period. Examples of less controversial directives, where there is less chance of unintended harms and more certainty in the operational environment, include:

1. Rules for robots sent into underground tunnels to find and help rescue miners
2. Rules for smart torpedoes sent to disable a ship
3. Rules for autonomous counter-cyber operations

Can the ACD framework also authorize more controversial objectives such as selecting a target and application of force as per AWS defined above? The argument of this paper is to say that iff AWS in a particular case are considered justified, then they ought to have an ACD to articulate this justification and to provide apprioriate scrutiny and accountability mechanisms. The paper does not promote any specific use of autonomous systems. Uncontroversial cases demonstrate that autonomy can be justified and that autonomy per se that is not the problem, but particular objectives, means of behaviour or actions that are problematic. Through the act of ACP and creation of an ACD the nuance of a scenario and permitted vs prohibited actions can be determined and authorized.

ACD must anticipate both strengths and weaknesses of autonomous systems. In the section following, autonomy command is considered against van Creveld's (1985) classic command and control (C2) framework.

*5.8.1 Command and Control*

How should Van Creveld's C2 functions be understood in autonomy command with advance control planning and ACD? Table 5.1 demonstrates the synergy and asymmetry between the traditional C2 framework and what would be required to plan command in advance.

*Table 5.1* van Creveld's functions of command and control considered against advance control planning and autonomous systems operating under an ACD

| VAN CREVELD (1985) FUNCTIONS OF COMMAND AND CONTROL | ADVANCE CONTROL PLANNING (ACP) | AUTONOMOUS SYSTEMS OPERATING UNDER AN ADVANCE CONTROL DIRECTIVE (ACD) |
|---|---|---|
| 1. COLLECTING INFORMATION ON OWN FORCES, THE ENEMY, THE WEATHER AND THE TERRAIN | Military training and operational readiness units prepare historical information on own forces, the enemy, likely weather and likely terrain. | Autonomous systems collect real-time information in theatre on own forces, the enemy, the weather and the terrain using sensors and on-board information processing, analysis and decision making; identifying any delta from baseline data provisioned in ACD. |
| 2. FINDING MEANS TO STORE, RETRIEVE, FILTER, CLASSIFY, DISTRIBUTE AND DISPLAY THE INFORMATION | Create advanced digital environments (digital twins) for simulation, integration and modelling of the battlespace with autonomous assets. Enable variable information flows and displays. | Autonomous systems use on-board information-processing components required in ACD. |
| 3. ASSESSING THE SITUATION | Expert team (comprising commander, legal advisors, simulation experts, decision scientists, technology experts and operational experts) assesses scenarios in simulated real-time, fast-tempo and slow-tempo scenarios in training and providing decision parameters for autonomous systems for real time operations without human intervention. | Autonomous systems assess the situation using sensor data and compare against scenarios described in ACD. |

| | | | |
|---|---|---|---|
| 4. | **ANALYZING OBJECTIVES AND FINDING ALTERNATIVE MEANS FOR ACHIEVING THEM** | Analyzing objectives and alternate means to achieve them in simulated real-time, fast-tempo and slow-tempo scenarios in training and providing action parameters for autonomous systems for real-time operations without human intervention. | Executing pre-determined means to achieve objectives from ACD based on updated situation data and reasoning from step 3 and likelihood of means to succeed in objectives. |
| 5. | **MAKING A DECISION** | Making decisions in simulated real-time, fast-tempo and slow-tempo scenarios in training and providing decision options and trigger conditions for autonomous systems for real-time operations without human intervention. | Make decisions in accordance with ACD-aligned with updated environmental or internal state changes (e.g. unexpected loss of fuel reduces flight range). |
| 6. | **PLANNING BASED ON THE DECISION** | Planning based on the decision in simulated real-time, fast-tempo and slow-tempo scenarios in training and providing sequence of actions required for autonomous systems to carry out the decision for real-time operations without human intervention. | Executing pre-determined sequence of actions as per ACD and outcomes of steps 3-5. Advance Control Planning may be updated with learnings from outcomes of the decision in 5 and new simulations created to plan the ACD for the next operation. |
| 7. | **WRITING AND TRANSMITTING ORDERS AS WELL AS VERIFYING THEIR ARRIVAL AND PROPER UNDERSTANDING BY THE RECIPIENTS** | Writing and transmitting orders in simulated scenarios as well as verifying their arrival and proper understanding by autonomous systems. | Execute pre-determined orders within parameters specified in ACD. |

| | | |
|---|---|---|
| 8. **MONITORING THE EXECUTION VIA FEEDBACK, AT WHICH THE PROCESS REPEATS ITSELF** | Monitoring the execution via feedback from simulated fast tempo and slow tempo scenarios in training. Use feedback to adjust parameters of operation of autonomous systems. | Monitoring and recording actions and supply record to commander upon recommencement of communications. |

In advance control planning, commanders should work out how to overcome the limitations of the technologies they have available to optimize their usefulness in the operational context, rather than lamenting limitations and vulnerabilities. Through the planning process, individual risk profiles of commanders will change the parameters of what autonomous systems can do when discharged to operate. This planning phase will also involve a different approach to writing orders for humans. The communication preferences, autonomy and agency of human subordinates calls for orders that allow human interpretation to 'get the job done' (Storr, 2009) in a way that autonomy command is likely to be more analytical; detailing many more options and authorising decision paths based on a wide range of actions and risks in diverse scenarios. Orders for autonomous systems might be conservative in some situations and more aggressive in others depending on the mission. For example, the value of ISR close to an enemy captured by a small autonomous UAV may be worth the platform itself being shot down. The same risk may be less likely to occur if a human asset were tasked to undertake ISR. The resultant ACD becomes a legal document in a similar manner to an advance care directive in medical contexts.

*5.8.2 The military mind*

What kind of military mind is needed to command autonomous systems? Clausewitz et al. (1976) considered intellect and temperament as critical. Chambers et al. (2017) summarize these as:

C i.   Courage in the face of danger
C ii.  Courage to accept responsibility for the effects of commander actions on people

- C iii.     Sensitive and discriminating judgment to manage information uncertainty, misinformation, miscommunication, and confusion of war
- C iv.     Resilience of the mind to emerge unscathed by the demands placed on it by uncertainty friction and danger
- C v.     Presence of mind to deal with the unexpected
- C vi.     The will to overcome the resistance from within the commander's organisation as the demands of war begin to affect their soldiers
- C vii.     Energy to remain staunch and endure prolonged resistance
- C viii.     Self-control

Does autonomy command require this same set of intellect and temperament? Overall, the intellect would be critical, but the Clausewitzian temperament may not be required. Item, C i. would seem unnecessary as the commander would not be in danger; C ii. remains critical, as does C iii.; and because planning would be done in a deliberate and slowed-down process during advance control planning, the mind of the autonomy commander discussed in C iv. may not have to be particularly resilient. The effect of slowing down time during advance control planning is to relieve autonomy command from some of the burden of real-time decision-making and improve training for C v. The strain of resistance C vi., the energy to endure C vii., and C viii self-control are all diminished in autonomy command. What might a new list look like? Some of what is needed are innate qualities of commanders:

- A i.     Courage to accept responsibility for the actions of autonomous systems
- A ii.     Openness to multi-disciplinary expertise and authority
- A iii.     Sensitive and discriminating judgment to manage information uncertainty, misinformation, miscommunication and confusion of war
- A iv.     Mental agility to consider future complex actions and deliberate between them
- A v.     Persistent temperament with attention to detail to endure many hours in simulation in advance control planning

Some of what autonomy command requires are skills that can be trained, including:

S i.     Expertise in military history, ethics, strategy and leadership

S ii.    Operational experience

S iii.   Probabilistic reasoning

S iv.   Systems thinking

S v.    Risk analysis

How will this work operationally? Will autonomy commanders first create an ACD for their autonomous assets and then go on to be real time commanders in theatre where those assets and conventional assets including people and platforms are deployed? Or are separate commanders chosen to operate in theatre—some traditional, some commanding autonomous assets? I expect the answer depends on the mission, the sort of autonomous systems deployed and the human capabilities available. Just as in the classical hierarchy of command, some autonomy commanders may be responsible for specific capabilities, e.g. setting up perimeter drone countermeasures in an operation and report to higher-level commanders. Perhaps as a general rule, the higher commander must understand the ACD of each autonomous capability under their command and accept ultimate responsibility for its deployment. If this is the case, then different training regimes will be required for different levels of the hierarchy. The commander in charge of preparing and signing the ACD will need an intimate knowledge of the capability. The senior leaders will need a suitable summary understanding of key decision points and mission parameters of the system prepared within the ACD.

The autonomy commander may be quite different to a traditional commander in temperament. Both will share strong commitment to excellence and moral courage, but diverse with regard to the way they thrive cognitively. The autonomy commander thrives in taking their time, focusing on details and hypothetical scenarios, tweaking risk parameters, consultation and collaboration, and documenting decision trees and uncertain branching futures as well as providing clear operational guidance and safety information. Whereas the Clausewitzian commander thrives cognitively in real time, in action within the tempo of battle, making decisions under uncertainty with dynamic information flows working with other people in real time.

While future commanders will still need to make decisions in real time with dynamic information displays, there will increasingly be opportunities to task autonomous assets to carry out commander's intent through an ACD that provides subtle and responsive mission parameters on the assets. Thus, it is increasingly important to know not just when AI is brittle and likely to make a mistake, but also when and how humans are likely to fall short. In some cases, humans will have insufficient training to manage the tasks they are given (or there was no training that could have prepared them for a unique situation). Humans may be eminently qualified but fatigued or multi-tasked such that they are unable to focus on the decision making required of them. ACD can be created and prepared in advance, anticipating a variety of circumstances that would justify the deployment of autonomous assets including for battlespace advantage (e.g. surprise, distraction, mass effects, precision fires etc…), sustaining the attention of an adversary, protection, intelligence, surveillance and reconnaissance (ISR) and so forth. AI can be trained in simulation on combinations of tasks that autonomous assets could be directed to undertake, to reduce the cognitive load on commanders and allow them to focus on tasks for which their capabilities are best utilized.

Take the challenge of time-critical decision making under uncertainty. Time-critical decision making is very demanding for decision makers. Commanders must consider risk, ambiguity, time and losses. If commanders take more time to make a decision pending learning new information that changes their perception of risk, there may be greater losses. In many military decisions there are no good outcomes, only choices between bad outcomes including unintended harms and loss of life and human decision makers must be protected from moral injury by the state (Enemark, 2019). To prevent moral injury, the state should not put commanders into decision making conditions where they lack agency and operate in unjust conflicts using unlawful means or methods of warfare. If a war is just and commanders are equipped to have agency and autonomy over decisions and are able to act lawfully, then the risk of moral injury is greatly reduced.

However, communicating uncertainty in risk descriptions to commanders may lead to delays in the decision-making process, due to aversion to ambiguity and other forms of uncertainty (Rydmark et al., 2021). Presenting uncertainty in a risk description in the form of a range is one way to ensure the decision maker understands the strength of knowledge. The width of the range informs the decision maker of the estimated probability for an event, so that the decision-maker can take this information into consideration when making decisions. But, decision-makers need to know not just range, but also the underlying distribution of the data that influences the probability. Researchers have found that specific training in processing uncertain information outside of a time critical decision can help humans make faster decisions within real time (Rydmark et al., 2021). Another form of uncertainty comes from context ambiguity—needing to know or define a context of action[8]. Agents may need indicator of likelihood of being in a particular context or contexts with perhaps a topological map or vectors or how different contexts are related. Advance control planning would involve probabilistic reasoning capability assessments (Stanovich et al., 2016) and training for commanders so that they were both better at battlespace decision making as well as authorising behaviours and decision making of autonomous agents (who are also conducting operations under uncertainty).

What should an autonomy commander include in an ACD to ensure moral and legal responsibilities? Nations should develop responsible AI frameworks for ACD to abide by (see policies, strategies and methods from USA, UK, Australia and NATO: Department of Defense Responsible AI Working Council, 2022; Devitt et al., 2021; Ministry of Defence, 2022; NATO, 2021). The ACP process ought to ensure that risks are managed and operators understand system behaviours and capabilities against guidance documents.

Advance control planning involves important reasoning and decision making about future command. It is an iterative process in the lead up to deployment where significant stressors to command are foreseeable including: fatigue, complexity, high tempo decision making, rapidly deteriorating situational awareness and unforeseen factors affect decision making. Plans should align with the commander's goals, risk preferences, personal values and other factors of individual autonomy provisioned in IHL to retain a human-centred, and humanity-centred approach to the use of autonomous technologies in armed conflict' and to preserve human control.

*5.8.3 Advance control directive*

The following are indicative contents within an ACD, the specifics of which must be determined by each nation State:

1. Description of the capability in general terms (e.g. weight, mobility, fuel, range, speed, sensors, comms, noise, vulnerabilities, redundancies)

2. Operational context assurances (e.g. terrain, weather conditions, team requirements as well as contextual vulnerabilities)
3. Mission types and parameters authorized within this directive
4. Governance frameworks including legal obligations
5. Commander goals, risk preferences, values and direction
6. Responsibility lines for autonomous functions designed by the manufacturer and human-directed behaviours authorized by the commander. Additional responsibilities as required, e.g. subordinate responsibilities if collocated with autonomous assets on the battlefield including witnessing autonomous systems behaviours and intervening as appropriate
7. Safety processes and procedures
8. Damaged platform management and retrieval requirements
9. Records functions describing internal processes and external environment for after action review

*5.8.4 Case study: Australia*

Australia's systems of control for autonomous weapons systems (Commonwealth of Australia, 2019) lists considerations that could be referenced and/or included in an ACD[9]:

1. Legal and policy foundations and design and development are guided by government direction and military end state. Technology from inception is constrained by laws, societal values, human rights and constitutional obligations. The scope of functionality and intended theatres of use are considered as well as coding security, force design, limitations and safeguards and human-machine interfacing.
2. Testing, evaluation and review through to acceptance, training and certification involves assurance and compliancy testing of systems and sub-systems including verification of software, reliability performance evaluation, software verification, limitation identification and certification and Article 36 reviews. Proficiency and legal training is undertaken by personnel.
3. Pre-deployment top-down strategic guidance is provided from the government and the military. Weapons systems are chosen and the parameters of their operation decided.
4. The decision to deploy requires government approval, force composure planning as well as pre-deployment training and certification.

5. The decision to employ includes targeting permissions, operating parameters and procedures, rules of engagement, targeting directives and authorizations.

Any of these stages are subject to after-action evaluation of use of force and an auditable trail of decision-making must be kept (Commonwealth of Australia, 2019).

Supposing that many personnel contribute to the ACD, when do commanders 'step in' to the systems of control? Commanders are expected to understand Australia's policies and legal obligations as well as government direction. Commanders should also maintain briefings on technologies in development and given the opportunity to workshop function, theatres of use, force design as well as human-machine interfacing. Technology developers need to ensure the human-machine interface is co-developed with command—because the interface should reveal the extent of situational understanding, and effect on human cognition, awareness, attention, responsiveness and capacity of command to abide by legal and ethical obligations. Ideally militaries will begin to standardize interfaces for managing multiple autonomous assets based on empirical human factors research and end-user engagement (Endsley, 2023; Hancock et al., 2008). Standard interfaces will reduce the time required of individual commanders in the design stages as well as reduce total training hours required to achieve accreditation at testing and review as well as pre-deployment. Commanders' influence is imperative as technologies are developed so that they can contribute to limitations and safeguards of the technology well ahead of deployment of these weapons systems. However, commanders will need to have a firm grasp on the nature of AI in autonomy command.[10]

Autonomy Command and Artificial Intelligence

> *...the exercise of technopower by and through AI takes place in time and even configures time, in the sense that it shapes our experience of time and configures our stories, days, and lives…. AI has the power to define my time. And by means of data collection and data analysis, my story is configured in terms of the statistical categories and profiles made by AI. This happens not only at the individual level but also at the cultural and societal level: our time becomes the time of AI and AI shapes the narrative of our societies.* (Coeckelbergh, 2022)

# Meaningful Human Command: Advance Control Directives as a Method to Enable Moral and Legal Responsibility for Autonomous weapons systems

Increasingly, militaries are coming to terms with how to integrate AI into battlespace decision making—augmenting what humans are good at and offloading what humans are bad at to computerized functions aiming to accelerate the pace and optimization of decision-making. Battlespace representations are complex, ambiguous and uncertain—a simulated 'fog of war'.

Battlespace information and predictive analytics based on machine learning (ML) are inductive and probabilistic. For example, a sensor on a UGV trained using an ML algorithm may report 95% confidence that an object in front of it is an adversary's truck, but that confidence relates only to conformance with the objective function of the model, not necessarily the real world[11]. In some contexts (say normal lighting conditions, expected obstacles, close fidelity to the training data set), the 95% confidence is well placed, and a commander could use the confident testimony of the ML to believe that there is an adversary's truck before the UGV. However, in an adversarial environment, the sensor might be 95% confident that there is a truck, but, actually there are no trucks, only cardboard cut outs of wheels designed specifically to spoof the sensor. Or perhaps conditions are foggy and the ML is misinterpreting all vehicles as trucks. Battlespace AI will need to be appropriately assured for use by commanders so that humans responsible for decisions are aware of when the system is likely to be giving an untrustworthy reading[12]. Is this feasible? After all it can be very difficult in advance to know when a system is behaving in an untrustworthy fashion. Indeed, trying to solve the issue of AI being aware of when it is operating in an untrustworthy fashion has been likened to the potentially intractable or uncomputable 'relevance' or frame problem in cognitive science (Sperber & Wilson, 1995; Wilson & Sperber, 2002; Xu & Wang, 2012). When humans do not understand communications from machines, the consequences can be dire—see the challenges of situational awareness on Air France 447 (Salmon et al., 2016). These questions reinforce the importance of including human factors and ergonomics experts in the design, development and testing of human-autonomy systems.

The assurance process will verify and validate the AI across a wide range of operational conditions and then humans working with AI will need also to be assured in simulation and training exercises in a variety of conditions likely to challenge decision making.

Information integration will be an important method to triangulate on reality and different artificial intelligence methods, such as Bayesian inference will help wrangle competing hypotheses regarding the situation. As an example, take the Waymo cars that have lidar, radar and cameras to offset errors during self-driving on San Francisco roads. Bayesian methods set a prior probability for the state of the world and then adjust their posterior probability based on the evidence received and the likelihood that the world is the way it is hypothesized based on that evidence. In an ideal world, Bayesian inference optimizes commander beliefs by helping determine which hypothesis is most likely given the data. Bayesian-driven AI can help in adversarial circumstances so long as independent evidence sources are used. However, it is possible for a creative adversary to seed disinformation across multiple types of data to systemically trick Bayesian systems. Military autonomous platforms will need to recognize that individual AI-driven sensors are brittle and combine their outputs in appropriate ways to make effective battlespace information environments. But, as Clausewitz presciently identified, even the best prepared militaries will be constantly challenged in the reality of war. Systems will be targeted by adversaries seeking to confuse commanders and commanders must remain vigilant to disinformation, misinformation and uncertainty. Regular and specific training in will be needed to keep up to date with what technology can do and likewise what adversaries can newly do.

In the far future, autonomous assets may be able to be tasked with an AI built on an accepted C2 common language ontology (Curts & Campbell, 2005; Kim et al., 2021; Smith et al., 2009). Some of the reports of artists 'commanding' Open AI's DALL-E to produce AI art in 2022 gives some insight into how the future of autonomy command might be structured. For example, the design lead at Cosmopolitan gave DALL-E hundreds of phrases to elicit an image of a female astronaut that suited the aesthetic of the magazine (karenxcheng, 2022). In the end the phrasing that produced suitable art was:

> A wide angle shot from below of a female astronaut with an athletic feminine body walking with swagger towards camera on mars in an infinite universe, synthwave digital art.

Elements that helped the AI understand the human requirement included semantic keys for functions (type of angle), concepts (e.g. athletic), actions (e.g. walking with swagger towards camera), background (e.g. infinite universe) and art style (e.g. synthwave).

S.K. Devitt

Elements that might prompt an AI assist with aligning with commander's intent include task type (e.g. planning, situational awareness, assessing), objects of interest (e.g. geofenced regions by GPS coordinates, lawful targets, protected objects), objectives (e.g. secure a building), constraints (0% harms to civilians) and so forth.

However, there are significant risks if the future of AI-assisted command is based on large transformer models containing the history of human bias and likely to reinforce and amplify bias, unfairness and unjust decision-making (Bender et al., 2021). The kinds of experts employed to manage the incorporation of AI within military decision making is crucial. Technical experts as well as ethical, safety and legal experts must be engaged. Commanders must be qualified and skeptical of AI when appropriate neither overtrusting or undertrusting their outputs.

## 5.9 The simulated battlespace

The potential of building ACD depends on some practical matters. In what sort of environment can all the considerations discussed so far be incorporated such that the relevant documentation can be prepared in advance of operations. I present the case study of modelling the fog of war in a simulated battlespace and suggest that militaries already have (or are building) some of the relevant digital architecture to improve developing and testing the performance of weapons systems. The framework discussed in this chapter can be incorporated into the wargaming and digital simulations of modern militaries. Combined with human factors assessments, operator and commander training, digital environments can perform multiple tasks including the preparation and finalization of ACDs to be approved and signed off by commanding officers ahead of operations. While the first ACD might be laborious the first time a weapons system is employed (in the same way that an article 36 can be), militaries would develop a library of documentation for a capability and a process to systematically update existing digital test, evaluation, verification and validation (TEVV) infrastructure with new performance and simulation data.

### 5.9.1 Case study: modelling the fog of war

For commanders to be morally and legally responsible for decisions they must both understand the situation they are in and have a reasonable level of awareness regarding the reliability of sensors and comms. When information from sensors and communications is unreliable, then command responsibility is at risk. The case study in Box 5.2 provides an example of how awareness can be tested in a wargaming environment. Note, the example is not meant to fully express the concepts of ACD, but is to indicate that there already exist simulation environments where parameters can be turned on and off to alter the epistemic experience of humans in these environments and that a similar design approach may help developers create simulation environments suitable to test numerous scenarios that toggle parameters that affect attribution of moral and legal command responsibility.

> **Box 5.2 Modelling the fog of war**
>
> A Fog Identification and Manipulation Methodology (FIMM) in a C2 decision-making process to manipulate sensor platform and communication links within an Advanced Framework for Simulation, Integration, and Modelling (AFSIM) wargame.
>
> The FIMM increases information confusion with inaccurate sensor information, communication links are delayed or data corrupted (Tryhorn et al., 2021). Examples of AFSIM sensor object functions include: azimuth error sigma, elevation error sigma, hits to establish track, maintain track probability. Examples of AFSIM communication object functions include: propagation speed, transfer rate, packet loss time, channels, queue type/limit, purge interval and retransmit attempts and delays. When subordinates are subjected to inaccurate or delayed data, they execute unintended orders that may not match the commander's intent.
>
> The Fog Analysis Tool (FAT) is an AFSIM plugin that allows each component of current sensors and communication components of a loaded scenario to be altered for a different fog effect including for platforms, processor, weapon, sensors, comm, mover, fuel and weapon effects shown in FIMM. FIMM allows the analyst to help train an agent to overcome fog when subjected to various fog effects. FAT needs more development of software architecture and user utilisation requirements before it can enter a verification and validation process. But its ability to disrupt information flow and introduce noise to sensor readings gives a preview of the kinds of tools that could be used to identify and

> operationalise risks in simulation ahead of a mission to provide design guardrails on the use of systems including any autonomous deployment.
>
> (Tryhorn et al., 2021)

One might wonder whether it is reasonable for a single commander to be morally responsible for the future decisions of an autonomous fleet. After all, if decisions are made in advance, and time can be taken in the planning stage, perhaps a human team working together should collectively hold responsibility for decisions? Consider the case of autonomous vehicles and the automated driver software. Legislation being considered by Australia suggests that when the autopilot is engaged in a self-driving car, if an accident occurs, the company that created the software is liable for incidents that occur caused by automated driver error (National Transport Commission, 2022). If a human takes over driver responsibilities in this vehicle, then the human becomes liable for incidents caused by their error.

There are important differences between civilian cars abiding by rules of the road to assist the transportation of vehicles versus military platforms abiding by commander's intent in a conflict. Unlike civilian use cases, actions in war are political, often applying force to confirm political intent with human consequences including human harms. The binding of actions in conflict to specific humans for legal and moral responsibility is critical in ensuring legitimacy of a belligerent's actions and for those actions to be understood by the international community in terms of appropriateness, proportionality and so forth. The chain of command from machines to humans is imperative to protect.

## 5.10 Conclusion

Emerging technologies are rapidly changing battlefield decision making and evidence-based normative frameworks are needed to guide their design, development, and adoption— particularly with regards to the appropriate role of the human in relation to their deployment. Critics and militaries both agree that war must remain a human endeavour and the technologies and systems engaged by humans must defer to human intent and values. It is agreed that humans need to understand weapons systems, be able to influence their operation on the battlefield and be legally and morally responsible for their deployment.

Keegan (2004) when considering the relationship between information, decision making and success in battle argues that war is 'ultimately about doing, not thinking'. This chapter has attempted to outline a process by which the thinking of battle, at least when deploying autonomous assets, can be done ahead of time so that the 'doing' can occur quickly and effectively with legal and moral responsibility intact. Indeed, rather than commanders being inundated with more and more information from the battlespace, some of that information can be interpreted by autonomous units that are authorized to act under specific conditions, such as deploying countermeasures to protect blue forces against an oncoming attack.

The chapter has discussed how advanced control plans and directives might be constructed to ensure moral and legal responsibility of autonomous weapons systems. The hope is that this chapter will be relevant for critics and militaries facing a changing strategic and operational environments, particularly where lethal decision processes occur in congested real-time, high-tempo and slow-tempo operations preventing humans from engaging in deliberative thought or direct control within real time to intervene on a decision by an autonomous weapons system.

Advanced control directives may benefit commanders by increasing their autonomy and wellbeing, preventing moral injury (Griffin et al., 2019) after conflicts as actions should align with the values and priorities of humans responsible for actions. However, such new instruments will need supporting policy and a military cultural environment backed by knowledgeable AI and autonomy governance.[13]

Meaningful Human Command: Advance Control Directives as a Method to Enable Moral and Legal Responsibility for Autonomous weapons systemsLuckett, T., Bhattarai, P., Phillips, J., Agar, M., Currow, D., Krastev, Y., & Davidson, P. M. (2015). Advance care planning in 21st century Australia: a systematic review and appraisal of online advance care directive templates against national framework criteria. *Australian Health Review*, *39*(5), 552-560.

Maan, A. K., & Cobaugh, P. (2018). *Introduction to Narrative Warfare: A Primer and Study Guide*. Narrative Strategies LLC.

McKenzie, S. (2020). When is a Ship a Ship? Use by State Armed Forces of Uncrewed Maritime Vehicles and the United Nations Convention on the Law of the Sea. *Melbourne Journal of International Law*, *21*(2), 373-402.

McKenzie, S. (2021). Autonomous technology and dynamic obligations: Uncrewed maritime vehicles and the regulation of maritime military surveillance in the exclusive economic zone. *Asian Journal of International Law*, *11*(1), 146-175.

Mettraux, G. (2009). *The law of command responsibility*. Oxford University Press.

Mill, J. S. (1860). [On Liberty](#) (2nd ed.). John W. Parker & Son.

Miller, A. (2021, September 21). AI Algorithms Deployed in Kill Chain Target Recognition. *Air & Space Forces Magazine*. https://airandspaceforces.com/ai-algorithms-deployed-in-kill-chain-target-recognition/

Ministry of Defence. (2022, June 15). Policy Paper: Defence Artificial Intelligence Strategy. https://www.gov.uk/government/publications/defence-artificial-intelligence-strategy/defence-artificial-intelligence-strategy#:~:text=This%20strategy%20sets%20out%20how,shape%20global%20AI%20developments%20to

Morrison, S. R. (2020). Advance directives/care planning: clear, simple, and wrong. *Journal of Palliative Medicine*, *23*(7), 878-879.

National Transport Commission. (2018). *Changing driving laws to support automated vehicles*. https://www.ntc.gov.au/transport-reform/ntc-projects/changing-driving-laws-support-AVs

NATO. (2021, October 22). *NATO releases first-ever strategy for Artificial Intelligence* https://www.nato.int/cps/en/natohq/news_187934.htmS.K. Devitt 41

**Notes**

[1] It is also important to have aggregative models to test what happens when multiple systems overlap (personal communication with Beth Cardier, 2022).

[2] Secretary of the Air Force Frank Kendall seeks AI capability to "Significantly reduce the manpower-intensive tasks of manually identifying targets—shortening the kill chain and accelerating the speed of decision-making" (Miller, 2021).

# Meaningful Human Command: Advance Control Directives as a Method to Enable Moral and Legal Responsibility for Autonomous weapons systems

[3] Comment made by a Bluebottle spokesperson during the Trusted Autonomous Systems, Autonomous Vessel Forum, 29 September 2022, Townsville, Australia.

[4] An example from Queensland Law "(1) A person is not criminally responsible for an act or omission if at the time of doing the act or making the omission the person is in such a state of mental disease or natural mental infirmity as to deprive the person of capacity to understand what the person is doing, or of capacity to control the person's actions, or of capacity to know that the person ought not to do the act or make the omission" .

[5] Thanks to Daniele Amoroso for providing this case law example.

[6] My thanks to Robert Bolia for suggesting Moltke and the relevance of his observation with the problem of autonomy command.

[7] Liivoja et al. (2022) conducted an analysis using the Vienna Convention on the Law of Treaties (VCLT) and found that there are no conclusive answers with regards to command of autonomous systems.

[8] Thank you to Beth Cardier for these insights on context uncertainty.

[9] Australia has not committed to the ACD concept. This case study is meant only as an aid to the research project, to provide some real-world solidity around which a hypothetical ACD might be considered.

[10] Including what kind of AI it is, e.g. expert system, ML, RL symbolic, Bayesian inference etc.

[11] Some may argue that this is also how human cognition works (thanks to Giulio Mecacci for this comment).

[12] Thank you to Giulio Mecacci for identifying this difficulty.

[13] I wish to thank Beth Cardier, Robert Bolia and anonymous reviewers of this chapter for their insightful comments.